\pdfoutput=1

\documentclass[11pt]{article}

\usepackage[final]{acl}

\usepackage{times}
\usepackage{latexsym}

\usepackage[T1]{fontenc}

\usepackage[utf8]{inputenc}

\usepackage{microtype}

\usepackage{inconsolata}

\usepackage{graphicx}

\usepackage{amsmath}
\usepackage{amsfonts}
\usepackage{booktabs}
\usepackage{multirow}
\usepackage{subcaption}
\usepackage{makecell}
\usepackage{color}
\usepackage{caption}
\usepackage{arydshln}
\usepackage{tabularx}
\usepackage{algorithm}
\usepackage{algorithmic}

\title{Making Large Language Models Better Reasoners with \\
 Orchestrated Streaming Experiences}

\author{
Xiangyang Liu \quad Junliang He \quad Xipeng Qiu\thanks{ {} Corresponding author.}\\
School of Computer Science, Fudan University \\
Shanghai Collaborative Innovation Center of Intelligent Visual Computing \\
\texttt{xyliu22@m.fudan.edu.cn}, \texttt{xpqiu@fudan.edu.cn}\\
}

\begin{document}
\maketitle
\begin{abstract}
Large language models (LLMs) can perform complex reasoning by generating intermediate thoughts under zero-shot or few-shot settings. However, zero-shot prompting always encounters low performance, and the superior performance of few-shot prompting hinges on the manual-crafted demonstrations. In this paper, we present \textbf{RoSE} (\textbf{R}easoning with \textbf{O}rchestrated  \textbf{S}treaming \textbf{E}xperiences), a general framework for solving reasoning tasks that can self-improve without complex external efforts. To enable RoSE, we describe an architecture that extends an LLM to store all answered questions and their thoughts in a streaming experience pool then orchestrates helpful questions from the pool to assist in answering new questions. To set up a question-aware orchestration mechanism, RoSE first calculates the similarity of each question in the pool with a new test question. Since the solution to each answered question is not always correct, RoSE will sort the questions according to their similarity with the new question, and then uniformly divide them into multiple buckets. It finally extracts one question from each bucket to make these extracted questions more diverse. To make these extracted questions help RoSE answer new questions as much as possible, we introduce two other attributes of uncertainty and complexity for each question. RoSE will preferentially select the questions with low uncertainty and high complexity from each bucket. We evaluate the versatility of RoSE in various reasoning tasks, LLMs, and CoT methods.
\end{abstract}

\section{Introduction}

Large language models (LLMs)~\cite{brown2020gpt3,thoppilan202lamda,chowdhery2022palm,Hoffmann2022chinchilla,ouyang2022instructgpt,zeng2023glm,touvron203llama,openai203gpt4,sun2024moss} have an emerged ability on performing various complex reasoning tasks. Recently, the chain-of-thought (CoT) prompting technique~\cite{wei2022cot} was proposed to have LLMs generate intermediate reasoning paths before generating the final answers. The prompting makes LLMs think deeply before giving an answer and further enhances the reasoning power of LLMs. Besides, the zero-shot CoT prompt~\cite{Kojima022zerocot} "Let's think step by step" also enhances the reasoning power of LLMs without any manual-crafting demonstrations. After the CoT prompting was proposed, more studies tried to manually design better prompts~\cite{zhou2023ltm,wang2023psp,yao2023tot} to further improve the performance of LLMs in reasoning. However, no matter how the prompts change, the goal is to have LLMs generate intermediate reasoning steps. 

Recent works such as ReAct~\cite{yao2023react}, Reflexion~\cite{shinn203reflexion}, REMEMBERER~\cite{zhang2023rememberer}, and ExpeL~\cite{zhao2023expel} were presented and have demonstrated the feasibility of autonomous agents that are built on top of an LLM core. These methods use LLMs to generate reasoning paths and “actions”. These "actions" can be used in API calls and executed in an environment. Besides, some golden feedback will be presented to LLMs during the reasoning process~\cite{shinn203reflexion,zhang2023rememberer} or labeled samples are needed to collect correct or false experiences~\cite{zhao2023expel}. Overall, these methods still require humans to carefully design some demonstrations and need golden feedback, labeled samples, or external tools to improve the reasoning performance of LLMs. 

We investigate how to improve the reasoning performance of LLMs in a more challenging streaming setting without any labeled data, pre-set unlabeled data, feedback signals, and other external help. Inspired by the observation that humans constantly do various exercises to construct a large experience pool in their minds and use the pool to help them quickly and better answer questions in exams, we present RoSE, a general framework for solving reasoning tasks with only streaming experiences. The greatest characteristic of RoSE is that it can self-improve by constantly collecting and orchestrating streaming experiences like humans. We build an experience pool for RoSE to store the answered questions and corresponding reasoning paths. We expect these questions can assist LLMs in answering new questions, and construct a novel experience orchestration mechanism to extract helpful questions from the pool for each new reasoning question. To achieve this, we consider three attributes for each question in the pool when orchestrating. First, the solution to each question may be incorrect. If we randomly select some answered questions as demonstrations, LLMs may directly copy the incorrect labels of these questions when they are similar to the questions to be answered. This phenomenon is also known as the \textit{copy effect}~\cite{lyu2023z-icl,zhang2033auto-cot}. To avoid this, we introduce \textbf{diversity} so that the extracted questions are distributed from the highest to lowest similarity to the question to be answered. Second, before a question is appended to the pool, we calculate \textbf{uncertainty} for it according to the outputs of LLMs. The lower the uncertainty, the more confident RoSE is about its prediction. We first filter questions with higher uncertainty in the pool. However, since the pool is a dynamic system, we also set the dynamic uncertainty threshold to only filter the questions with relatively higher uncertainty in a pool snapshot. Third, one intuition is that the more complex the question, the more it can help RoSE learn how to answer other questions~\cite{fu2023complexity}. Therefore, we introduce the \textbf{complexity} as the final attribute. After filtering the questions with high uncertainty, we select the most complex questions as the final demonstrations. 

We evaluate the versatility of RoSE on 9 reasoning tasks, 2 LLMs, and different CoT methods. Experimental results show that RoSE significantly improves the reasoning performance of LLMs. The analysis experiments verify the importance of each experience orchestration process and the stability of RoSE across various experimental settings. We summarize our contribution as follows:
\begin{itemize}
\item We present RoSE, a general framework for better solving reasoning tasks. We build a novel experience orchestration mechanism by introducing diversity, uncertainty, and complexity to extract more helpful questions to assist LLMs in answering new questions. RoSE can self-improve by constantly answering new questions without complex external effort.
\item We verify the versatility of RoSE on 9 reasoning tasks, 2 LLMs, and different CoT methods. Experimental results show that RoSE can significantly improve the reasoning performance of LLMs.
\item We conduct extensive further analyses and show that each component of RoSE contributes critically to the improvements and also verify the stability of RoSE across various experimental settings. Code is publicly available at \url{https://github.com/xyltt/RoSE}.
\end{itemize}

\section{Related Work}

\subsection{Chain-of-Thought Prompting}
\citet{wei2022cot} formally presented the CoT prompting in large language models. This technique elicits LLMs to generate a series of intermediate reasoning steps that lead to the final answer to a question using some manual-crafting demonstrations with reasoning steps, so we name it \textbf{Few-Shot-CoT}. \citet{Kojima022zerocot} presented that LLMs can also perform CoT reasoning when prompted by a "magic spell" of "Let's think step by step" without any other manual-crafting demonstrations, so we name it \textbf{Zero-Shot-CoT}. We categorize prompting methods as zero- and few-shot settings.

\paragraph{Zero-shot Setting} Some studies tried to first use zero-shot CoT prompting to obtain the reasoning chain for each unlabeled question and build a retrieval mechanism to retrieve some helpful questions to construct a few-shot prompt. For example, Auto-CoT~\cite{zhang2033auto-cot} uses the k-means clustering method to cluster all the test questions except the current question to be answered, then takes all the questions near each cluster center to construct a few-shot prompt using zero-shot CoT prompting. Plan-and-Solve prompting~\cite{wang2023psp} uses a different zero-shot CoT prompt to elicit LLMs to first decompose a question into sub-questions and then solve each sub-question.

\paragraph{Few-shot Setting} Few-shot CoT prompting achieves better performance by eliciting the CoT reasoning ability with effective manual demonstrations. However, designing suitable prompts for all test questions is difficult. Some recent studies mainly focus on manual-crafting more well-designed prompts instead of addressing this limitation. \citet{zhou2023ltm} and \citet{Khot2023decomp} presented similar CoT prompts to first decompose a complex question into multiple sub-questions and then solve them one by one. PoT~\cite{chen2022pot} uses a CoT prompt to elicit LLMs to generate text and programming language statements where the generated program can be executed by a program interpreter to get the final answer. \citet{fu2023complexity} presented a complexity-based few-shot CoT prompting method that uses more complex demonstrations (i.e., with more reasoning steps) to obtain better performance than a random few-shot CoT prompt. \citet{yao2023tot} presented a Tree-of-Thought (ToT) prompting method by considering multiple different reasoning paths and self-evaluating choices to decide the next course of action. MoT~\cite{li2023mot} obtains the reasoning paths for each unlabeled question using few-shot CoT prompting and filters the questions with low confidence. MemPrompt~\cite{madaan2022memprompt} also uses few-shot prompting to query LLMs and gathers the interaction histories with user feedback to concatenate with the original prompt. Besides, there are many retrieval-based in-context learning methods~\cite{luo2024icl_survey} that leverage existing databases and retrieval systems. Unlike these methods, RoSE puts more emphasis on the self-improvement of LLMs without any external data or feedback.

\subsection{Reasoning with Language Agents}
Some studies built agents to solve reasoning and decision-making tasks. ReAct~\cite{yao2023react} explores the use of LLMs to generate both reasoning traces and task-specific actions in an interleaved manner. Reflexion~\cite{shinn203reflexion} is an agent with memory and self-reflection and can be used to solve reasoning and decision-making tasks. ExpeL~\cite{zhao2023expel} is an agent that can learn from experiences and insights. However, it needs labeled data to construct experiences and insights. Compared with these agents, RoSE does not require external environments or feedback.


\begin{figure*}[t]
\centering
\includegraphics[width=1.99\columnwidth]{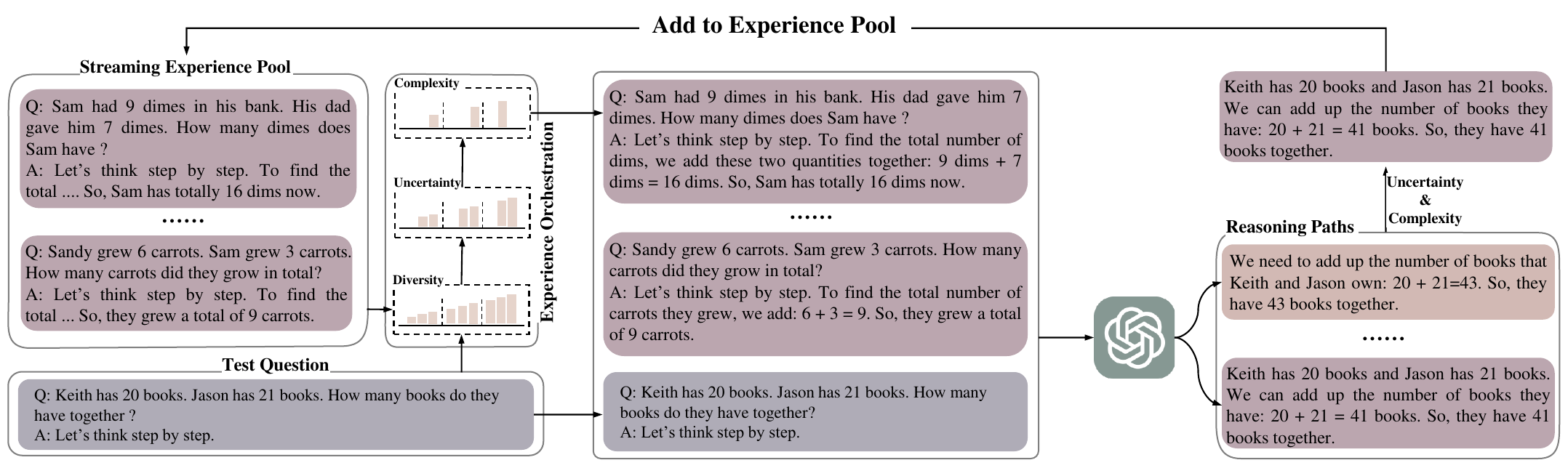}
\caption{The overview of RoSE}
\label{fg:overview}
\end{figure*}

\section{Methodology}

In this paper, we present RoSE, a framework for collecting and orchestrating streaming experiences to make LLMs self-improve in various reasoning tasks. Our setting is zero-shot (i.e., without any manual-crafting demonstrations) and streaming (i.e., test questions arrive one by one and there are no pre-set unlabeled questions). Figure~\ref{fg:overview} shows the overview of the proposed framework. RoSE incorporates a streaming experience pool to store the answered questions and their reasoning paths. RoSE will orchestrate the experiences using multiple attributes to extract helpful questions to assist itself in better answering new questions. We construct a novel experience orchestration mechanism for RoSE that considers the diversity, uncertainty, and complexity of questions. In this section, we introduce how RoSE collects streaming experiences and how it orchestrates the collected experiences.

\subsection{Streaming Experience Pool}
\label{sec:pool}
The streaming experience pool is a dynamic system to store the answered questions and their reasoning paths. After answering a new question, RoSE will store it and its reasoning path in the streaming experience pool. Each answered question has two attached attributes of uncertainty and complexity according to the predictions of RoSE. The two attributes will be regarded as important measures to filter collected experiences.

\begin{figure}[ht]
    \centering
    \includegraphics[width=\linewidth]{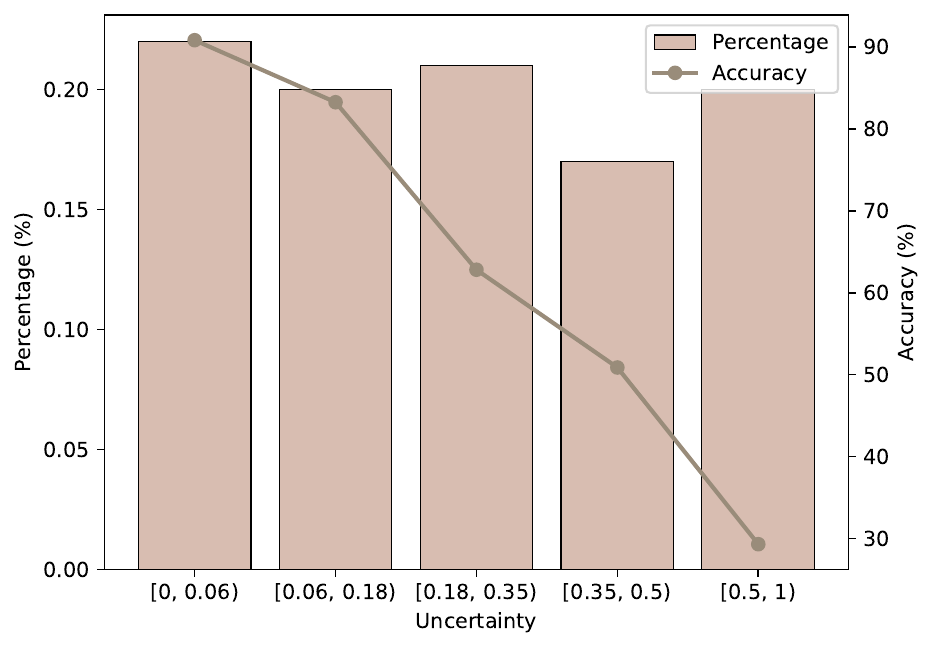}
    \caption{The relation between accuracy and the magnitude of uncertainty value on SVAMP dataset. We normalize the range of uncertainty to [0, 1].}
    \label{fig:uncertainty}
\end{figure}

\paragraph{Uncertainty} The uncertainty attribute indicates how confident RoSE is in answering a question. As shown in Figure~\ref{fig:uncertainty}, the lower the uncertainty, the more confident RoSE answers the question. RoSE will filter the questions in the experience pool with higher uncertainty to guarantee the correctness of extracted questions. To calculate uncertainty, we make LLMs generate multiple reasoning paths for each question. Each reasoning path has a corresponding predicted answer. Following~\citet{li2023mot}, We calculate an entropy to estimate uncertainty according to all predicted answers $\mathcal{A}$:
\begin{align}
    & \mathcal{A}^* = \rm Unique(\mathcal{A}), \\
    & p(a_i^*) = \sum\nolimits_{j=1}^{m} \mathbb{I}(a_i^* = a_j) / m, \\
    & u_{q_t} = -\sum\nolimits_{i=1}^{\left| \mathcal{A}^* \right|} p(a_i^*)\; {\rm log} \;p(a_i^*),
\end{align}

where $m$ is the number of reasoning paths and $\mathcal{A} = [a_1, a_2, ..., a_m]$ is the corresponding answers of each reasoning path for the test question $q_t$. $\mathcal{A}^* = [a_1^*,\;a_2^*,\;...]$ is the set of answers $\mathcal{A}$. $u_{q_t}$ represents the uncertainty of test question $q_t$ and the higher $u_{q_t}$ is, the more uncertain the LLM is about the question. 

\paragraph{Complexity} An intuition is that the more complex a question, the more it includes the details of the reasoning that can better teach LLMs how to reason. Therefore, we introduce the complexity attribute for each question as another important measure when filtering experiences. A natural idea is to use the average complexity of the reasoning paths to represent the complexity of a question. The higher the average path complexity, the more complex the question. For example, when a math word problem is more complex, it may require more columns of equations, resulting in more complex reasoning paths. Therefore, we measure the complexity of a question $q$ as follows:
\begin{align}
    & c_{q} = \sum\nolimits_{i=1}^{\left|\mathcal{R}^*\right|}{\rm CountSteps}(r_i)/\left|\mathcal{R}^*\right|,
\end{align}
where $\mathcal{R}^*$ is the set of reasoning paths corresponding to the most frequent predicted answer and $\rm CountSteps(\cdot)$ is a function to obtain the number of steps in a reasoning path $r$. Following~\citet{fu2023complexity}, we see a line as one reasoning step.

\paragraph{Experience Collection}
As just discussed, RoSE generates $m$ reasoning paths for each test question. However, we only select one reasoning path and add it to the streaming experience pool. To guarantee more reasoning details, we select the path with the most reasoning steps:
\begin{align}
    & r^* = \max (\mathcal{R}^*,\, {\rm key}={\rm CountSteps}).
\end{align}

Table~\ref{tab:experience_pool} depicts a demonstration of the collected experiences. RoSE will orchestrate these experiences to better assist itself in answering new questions.

\begin{table}[ht]
\resizebox{0.98\linewidth}{!}{
\begin{tabular}{ccccc}
\toprule
\textbf{Question} & \textbf{Rationale} & \textbf{Answer} & \textbf{Uncertainty} & \textbf{Complexity}  \\
\midrule
$q^1$ & $r^1$ & $a^1$ & $u^1$ & $c^1$  \\
$q^2$ & $r^2$ & $a^2$ & $u^2$ & $c^2$  \\
$q^3$ & $r^3$ & $a^3$ & $u^3$ & $c^3$  \\
\vdots & \vdots & \vdots & \vdots & \vdots \\
\bottomrule
\end{tabular}}
\caption{An example of the experiences stored in the experience pool.}
\label{tab:experience_pool}
\end{table}


\subsection{Experience Orchestration}
RoSE will orchestrate the collected experiences to assist itself in answering new questions. It first considers the diversity of experiences, and then filters useless questions using the attached attributes of uncertainty and complexity sequentially. Finally, it constructs a CoT prompt using the orchestrated experiences.

\paragraph{Diversity} Recent studies found that LLMs will directly copy the wrong labels from the ICL demonstrations~\cite{lyu2023z-icl} or be misled by the wrong predictions in demonstrations~\cite{zhang2033auto-cot} if the demonstrations in prompts are very similar to test questions. Therefore, some recently proposed methods~\cite{zhang2033auto-cot,li2023mot} consider diversity when constructing demonstrations using unlabeled questions. Different from these methods that use k-means clustering, we propose a question-aware approach to maintain diversity. Specifically, given a test question $q_t$ and the answered questions $(q^1, q^2, ..., q^j)$ in the experience pool, we first obtain their embedding representations using an off-shelf semantic embedder. Then we calculate the semantic similarity between the answered questions and the test question using their embedding representations. The answered questions are sorted from low to high semantic similarity and uniformly partitioned into $k$ buckets at the dimension of similarity, where $k$ is the number of demonstrations. The process of partitioning is summarized in Algorithm~\ref{alg:partiion}. RoSE will select one question in each bucket. This makes the selected questions distribute from low similarity to high similarity to the test question and guarantees the diversity of selected questions. We show that this can perform better than Auto-CoT which uses the k-means clustering method in the latter section.

\paragraph{Uncertainty-based Filtering} After partitioning the answered questions into $k$ buckets, RoSE will filter the answered questions with high uncertainty in each bucket. The streaming experience pool is a dynamic system and the uncertainty distribution among all buckets is different in different snapshots. Moreover, the uncertainty distribution is also different for different tasks. Therefore, a fixed filtering threshold does not necessarily work well for every bucket and we can not find an applicable threshold for each task. To ease the awkward situation, we propose to set a dynamic uncertainty threshold for each bucket to guarantee that RoSE only filters out the questions with relatively high uncertainty in each bucket and there are no empty buckets after filtering. Specifically, for each bucket, we adopt the $\lambda$ times of minimal uncertainty value in the bucket as the threshold and filter out the questions whose uncertainty is higher than the threshold:
\begin{align}
f(b_i) &= \{ q \in b_i \mid u_{q} <= \lambda \cdot u_i^{min} \}, \\
u_i^{min} &= \min\{q \in b_i \mid u_{q} \},
\end{align}
where $b_i$ indicates bucket $i$ and $u_i^{min}$ indicates the minimum uncertainty value of the bucket $i$.


\begin{algorithm}[t]
\caption{Partition}
\label{alg:partiion}
\begin{algorithmic}[1] 
\REQUIRE $q_t,\,\mathcal{Q}_a = [q^1,\, q^2,\, ..., q^j]$ and $k$\\
\STATE Calculate the similarity of each question pair $(q_t,\, q^1),\, ...,\, (q_t,\, q^j)$
\STATE Sort $q^1, \,q^2,\, ...,\, q^j$ through the magnitude of similarity 
\STATE Uniformly partition $\mathcal{Q}_a$ into $k$ buckets at the dimension of similarity, represented by $\mathcal{B} = [b_1,\, b_2,\, ...,\, b_k]$ 
\STATE Remove empty buckets in $\mathcal{B}$
\WHILE{$len(\mathcal{B}) < k$}
\STATE Select the bucket with the highest number of questions and uniformly partition it into 2 buckets.
\ENDWHILE
\STATE \textbf{return} $\mathcal{B}$
\end{algorithmic}
\end{algorithm}
   
\paragraph{Complexity-based Filtering} The final filtering is complexity-based. As mentioned before, the more complex a question, the more it includes the details of the reasoning that can better teach LLMs how to reason. Therefore, we select the question with the highest complexity from each bucket:
\begin{align}
q^i &= \max (b_i, {\rm key}=c_{q}).
\end{align}

\subsection{Inference} Given a test question $q_t$, RoSE orchestrates the experiences to extract $k$ experiences from the streaming experience pool and the unit of each experience is a triplet (question, rationale, answer). Finally, it answers the test question in the following manner:
\begin{align}
   & o_t = LLM(q^1, r^1, a^1, \,...,\,q^k, r^k, a^k,\,q_t) \\
   & r_t, a_t = {\rm ParseAnswer}(o_t)
\end{align}

\begin{table*}[t!]
\resizebox{\linewidth}{!}{
\begin{tabular}{lcccccccccc}
\toprule
\multirow{2}{*}{\textbf{Method}} & \multicolumn{6}{c}{\textbf{Arithmetic}}                       & \multicolumn{3}{c}{\textbf{Common Sense}} & \multirow{2}{*}{\textbf{AVG}}  \\ 
\cmidrule(r){2-7} \cmidrule(r){8-10}  & AddSub & AQuA & GSM8K & SingleEq & SingleOp & SVAMP & CSQA    & Strategy   & Date       \\ 
\midrule
\multicolumn{11}{c}{\textit{GPT-3.5-Turbo-16k-0613}} \\
\midrule
Zero-Shot-CoT        & 83.5   & 55.5 & 75.8  & 90.9     & 90.9     & 77.5  & 67.6    & 65.5      & 67.5      & 75.0    \\ 
Few-Shot-CoT         & 88.6   & 55.1 & 75.4  & 93.7     & 90.9     & 80.6  & 66.7    & 68.0       & 78.3        & 77.5     \\ 
Auto-CoT             & \textbf{91.4}  & 52.8 & 74.4  & 91.5     & 93.6     & 84.9  & 74.8    & 62.0           & 56.6    & 75.8   \\ 
\hdashline
Zero-Shot-CoT-SC     & 85.1   & 61.8 & 77.6  & 93.3     & 92.5     & 84.3  & 72.1    & 66.3         & 75.1    & 78.7    \\ 
Few-Shot-CoT-SC      & 89.1   & 58.7 & 82.0  & \textbf{94.5}     & 94.8     & 86.4  & 68.8    & 69.9         & 79.9    & 80.5    \\ 
Auto-CoT-SC          & 89.4   & 61.8 & 80.0  & 92.5     & 91.6     & 88.5  & \textbf{77.0}    & 63.9      & 78.0  & 80.3    \\ 
\hdashline
RoSE (Ours)       & 90.9   & \textbf{70.9} & \textbf{83.9}  & 92.2     & \textbf{95.6}     & \textbf{89.2}  & 67.8    & \textbf{71.3}    & \textbf{88.6}  & \textbf{83.4}     \\
\midrule
\multicolumn{11}{c}{\textit{LLaMA2-13B-Chat}} \\
\midrule
Zero-Shot-CoT      &  14.7 &	14.2 &	9.0 &	18.5 &	16.2 &	17.3 &	33.1 &	57.4 &	37.7 & 24.2    \\ 
Few-Shot-CoT         &   37.5 &	26.0 &	16.6 &	43.1 &	53.2 &	38.2 &	24.0 &	68.1 	& 	58.3  &	40.6      \\ 
Auto-CoT             &   58.5 &	22.4 &	35.9 &	69.5 &	81.0 &	38.2 &	61.7 &	63.0  	& 56.6 	 & 54.1   \\ 
\hdashline
Zero-Shot-CoT-SC     &    52.4 &	19.3 &	31.1 &	58.9 &	45.6 &	50.0 &	39.1 &	63.6 	&36.0 &	44.0   \\ 
Few-Shot-CoT-SC      &  57.5 &	26.8 &	31.4 &	62.6 &	70.5 &	57.7 &	26.1 &	68.0 	&	54.2  &	50.5  \\ 
Auto-CoT-SC          &   69.9 &	24.4 	& 48.1	& 79.9 &	86.3 &	63.5 &	54.7 &	60.3  &	55.0   & 60.2   \\ 
\hdashline
RoSE (Ours)             &   \textbf{79.5} &	\textbf{31.5} &	\textbf{50.2} &	\textbf{81.3} &	\textbf{89.5} &	\textbf{64.3}  & \textbf{62.2}	&	\textbf{69.4} &	\textbf{63.7}  &  \textbf{65.7}  \\
\bottomrule
\end{tabular}}
\caption{Main results for RoSE. "SC" represents self-consistency~\cite{wang2023sc}.}
\label{tab:result_wo_stm}
\end{table*}

\section{Experiments}
We conduct a series of experiments to compare the proposed RoSE with existing approaches on various reasoning tasks. We find that RoSE robustly improves reasoning capability in different experimental settings and each process of orchestrating experiences is important.

\subsection{Experimental Settings}

\paragraph{Models} We conduct all the main experiments on two large language models including \texttt{gpt-3.5-turbo-16k-0613} and \texttt{LLaMA2-13B-Chat}~\cite{TOUVRON2023llama2}. For the semantic embedder, we use \texttt{all-mpnet-base-v2}~\cite{reimers2013sbert}. To save the cost, we conduct the most analysis experiments on \texttt{LLaMA2-13B-Chat} unless otherwise specified.
\paragraph{Tasks and Datasets} We evaluate RoSE on 9 reasoning tasks. By default, we use the test split for all datasets if the labels are available for evaluation. For StrategyQA, we randomly select 800 samples from test sets to be evaluated. The detailed statistics of each dataset can be found in Appendix~\ref{sec:dataset}.

\paragraph{Method Comparison} Since we mainly focus on the streaming setting without any labeled data and pre-set unlabeled data, we compare RoSE with Zero-Shot-CoT, Few-Shot-CoT, and Auto-CoT. To make a more fair comparison, we also compare the self-consistency~\cite{wang2023sc} version of these baseline methods. For Auto-CoT, we also adopt the same streaming setting as RoSE.

\paragraph{Implementation Settings} We use the temperature $T = 1.0$ when generating diverse reasoning paths and 20 reasoning paths will be generated for each question. We adopt $\lambda=1.2$ times of minimal uncertainty value in each bucket as the threshold unless otherwise specified. For the methods that do not need to generate multiple diverse reasoning paths, we use the temperature $T = 0$. We conducted all experiments on 8 Nvidia A100 GPUs.

\subsection{Main Results}


\begin{figure}[ht]
    \centering
    \includegraphics[width=\linewidth]{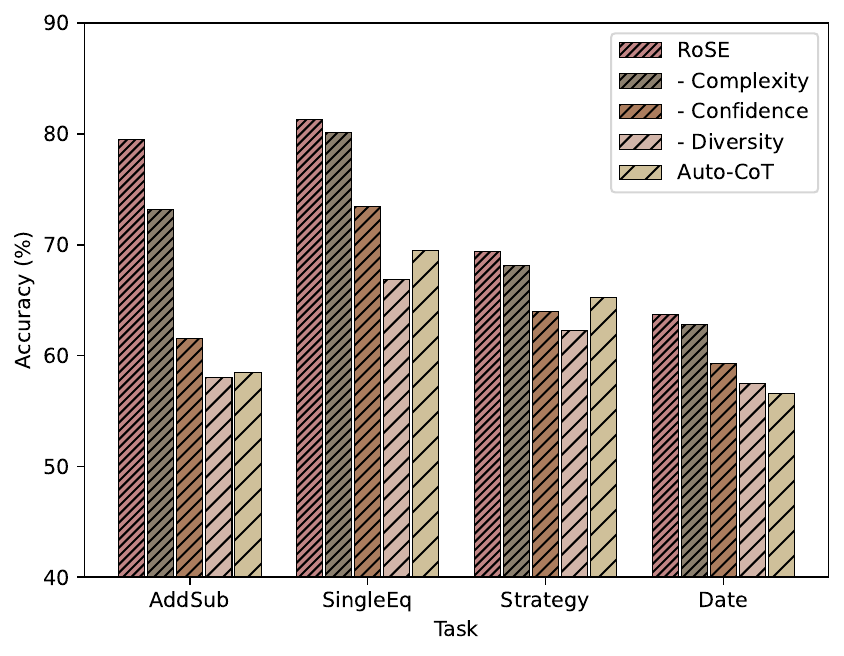}
    \caption{The impact of each orchestration process.}
    \label{fig:process}
\end{figure}

\label{analysis}

According to the comparison results in Table~\ref{tab:result_wo_stm}, RoSE performs better than all baselines overall. For the results on GPT-3.5-Turbo, RoSE exceeds Zero-Shot-CoT and Few-Shot-CoT by 8.4 and 5.9 points respectively and exceeds Zero-Shot-CoT-SC and Few-Shot-CoT-SC by 4.7 and 2.9 points respectively. This directly demonstrates that RoSE can self-improve by only the collected streaming experiences. While Few-Shot-CoT prompting uses demonstrations with human annotations, these demonstrations do not necessarily work for all test questions. However, RoSE has a big advantage over Few-Shot-CoT prompting by orchestrating helpful demonstrations from the experience pool for each test question. RoSE also shows significant improvements to Auto-CoT that only considers the diversity of demonstrations, and this indicates the importance of our proposed well-designed experience orchestration mechanism. 

Compared to GPT-3.5-Turbo, LLaMA2-13B-Chat has a big capacity gap on all reasoning tasks. However, RoSE also performs better than all baseline methods overall on LLaMA2-13B-Chat model and the improvement becomes larger than it on GPT-3.5-Turbo. After equipping with RoSE, the performance of LLaMA2-13B-Chat on multiple tasks approaches GPT-3.5-Turbo, such as SingleEq and StrategyQA.

\begin{table}[ht]
\resizebox{0.98\columnwidth}{!}{
\begin{tabular}{lcccccc}
\toprule
\multicolumn{1}{c}{} &\multicolumn{3}{c}{\textbf{Dynamic Threshold}} &\multicolumn{3}{c}{\textbf{Fixed Threshold}} \\
        \cmidrule(r){2-4}  \cmidrule(r){5-7} & 1.2 & 1.4 & 1.6& 0.6 & 1.2 & 1.8  \\
\midrule
AddSub      & \textbf{79.5} & 78.2  & 77.7 & 69.4 & 73.6  &  73.4    \\
SingleEq      & \textbf{81.3} & 80.9  & 79.7 & 79.9 & 81.1  &  79.8      \\
Strategy      & \textbf{69.4} & 69.3  & 68.1 & 67.1 & 68.9  & 68.2       \\
Date      & \textbf{63.7} & 61.5  & 62.1 & 57.7 & 60.9  &  60.1      \\
\bottomrule
\end{tabular}}
\caption{The impact of uncertainty threshold.}
\label{tab:ablation_para}
\end{table}

\subsection{Analyses}

\paragraph{The Effect of Each Orchestration Process} To better understand the contribution of each experience orchestration process, we conduct comprehensive ablation studies on four tasks. The ablation results are shown in Figure~\ref{fig:process}. We can observe that through the gradual orchestration process from diversity to uncertainty to complexity, the overall performance of RoSE on four datasets is gradually improved. This means that each process we propose increases the helpfulness of the extracted experiences in answering new questions. RoSE that takes uncertainty into account shows a jump in performance compared to the one that does not because the former generates multiple reasoning paths for each question and makes a majority vote among all predicted answers. Besides, RoSE which only considers diversity performs better than Auto-CoT overall. This represents the proposed question-aware diversity maintaining method is superior to the methods that the k-means clustering method used by Auto-CoT.   

\begin{figure}[ht]
    \centering
    \includegraphics[width=\linewidth]{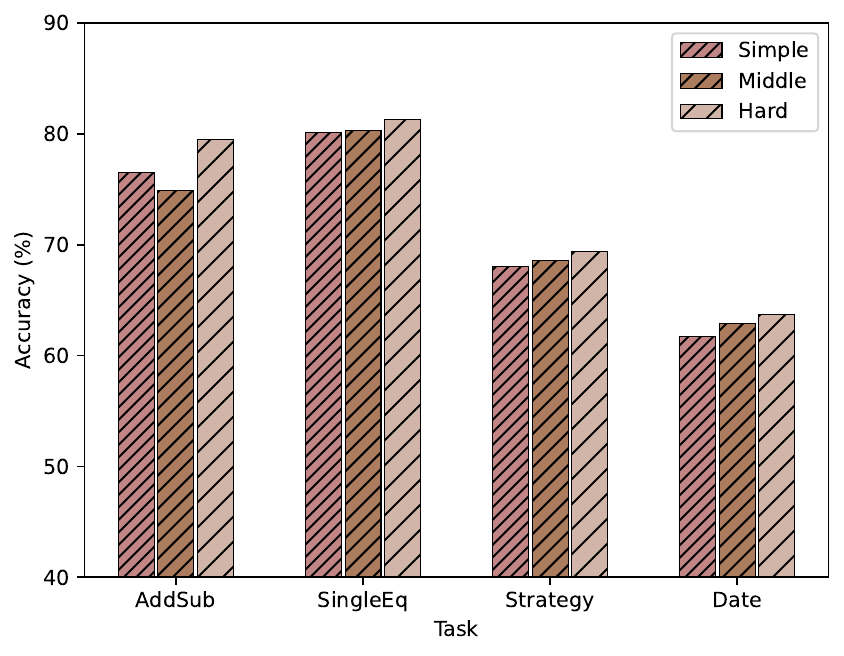}
    \caption{The impact of complexity.}
    \label{fig:ablation_complexity}
\end{figure}

\begin{table}[ht]
\resizebox{0.98\columnwidth}{!}{\begin{tabular}{lccccc}
\toprule
\multicolumn{1}{l}{\textbf{Method}} & \textbf{AddSub} & \textbf{SingleEq} & \textbf{Strategy} & \textbf{Date} & \textbf{AVG} \\
\midrule
\multicolumn{6}{c}{\textit{Temperature = 0.8}}  \\                        \midrule
Zero-Shot-CoT-SC                    & 50.1            & 57.9              & 61.6              & 36.0            & 51.4         \\
Few-Shot-CoT-SC                     & 54.4            & 59.8              & 67.3              & 53.1          & 58.7         \\
Auto-CoT-SC                         & 64.1            & 76.9              & 63.3              & 51.3          & 63.9         \\
RoSE (Ours)                         & \textbf{75.4}            & \textbf{80.3}              & \textbf{68.4}              & \textbf{63.4}          & \textbf{71.9}         \\
\midrule
\multicolumn{6}{c}{\textit{Temperature = 1.2}}                                                                               \\
\midrule
Zero-Shot-CoT-SC                    & 54.4            & 59.6              & 64.3              & 34.4          & 53.2         \\
Few-Shot-CoT-SC                     & 62.0             & 65.2              & 68.2              & 55.3          & 62.7         \\
Auto-CoT-SC                         & 73.1            & 77.2              & 60.9              & 57.8          & 67.3         \\
RoSE (Ours)                         & \textbf{80.3}            & \textbf{81.9}              & \textbf{69.8}             & \textbf{65.9}          & \textbf{74.5}      \\
\bottomrule
\end{tabular}}
\caption{The results on different temperatures.}
\label{tab:ablation_temperature}
\end{table}

\begin{table}[ht]
\resizebox{0.98\columnwidth}{!}{\begin{tabular}{lccccc}
\toprule
\multicolumn{1}{l}{\textbf{Method}} & \textbf{AddSub} & \textbf{SingleEq} & \textbf{Strategy} & \textbf{Date} & \textbf{AVG}  \\
\midrule
\multicolumn{6}{c}{\textit{Resoning Paths = 10}}                                                                              \\
\midrule
Zero-Shot-CoT-SC                    & 49.4            & 56.7              & 59.2              & 33.3          & 49.7          \\
Few-Shot-CoT-SC                     & 57.0            & 58.7              & 63.3              & 53.9          & 58.2          \\
Auto-CoT-SC                         & 69.0            & 74.9              & 57.3              & 51.3          & 63.1          \\
RoSE (Ours)                         & \textbf{77.2}   & \textbf{76.6}     & \textbf{67.8}     & \textbf{63.7} & \textbf{71.3} \\
\midrule
\multicolumn{6}{c}{\textit{Resoning Paths = 15}}                                                                              \\
\midrule
Zero-Shot-CoT-SC                    & 51.1            & 57.7              & 61.8              & 35.8          & 51.6          \\
Few-Shot-CoT-SC                     & 59.5            & 60.0              & 66.2              & 52.6          & 59.6          \\
Auto-CoT-SC                         & 73.9            & 76.3              & 58.9              & 53.6          & 65.7          \\
RoSE (Ours)                         & \textbf{77.9}   & \textbf{79.4}     & \textbf{69.1}     & \textbf{62.3} & \textbf{72.2} \\
\bottomrule
\end{tabular}}
\caption{The results on different numbers of reasoning paths.}
\label{tab:ablation_path}
\end{table}



\paragraph{The Impact of Different Uncertainty Thresholds} As shown in Table~\ref{tab:ablation_para}, we compare the performance of RoSE with different uncertainty thresholds. As introduced in the previous section, we adopt $\lambda$ times the minimal value of uncertainty in a bucket as the uncertainty threshold of the bucket. We first compare the performance of RoSE when adopting different values for $\lambda$. We find that the value of lambda values should not be too large, or RoSE may retrieve ones with high uncertainty, resulting in lower performance. Moreover, we also evaluate the performance of RoSE with a fixed uncertainty threshold for each bucket. Using a fixed threshold leads to lower performance than RoSE with a dynamic uncertainty threshold. This represents selecting a suitable fixed threshold for different buckets is difficult and also proves that the adopted dynamic threshold is robust.

\paragraph{The Impact of Different Complexity Thresholds} As shown in Figure~\ref{fig:ablation_complexity}, we also compare the performance of selecting the questions with different complexity and find that the more complex the extracted questions, the more helpful they are. This is also consistent with our initial intuition mentioned in Sec~\ref{sec:pool}, that the more complex a question, the more it includes the details of the reasoning that can better teach LLMs how to reason.

\paragraph{Results on Different Temperature Values} In this section, we evaluate RoSE under different temperature values. Table~\ref{tab:ablation_temperature} shows the results. We observe that RoSE consistently outperforms baseline methods across different temperature values, which shows the stability of RoSE. Besides, RoSE performs worse when adopting a temperature of 0.8 than a temperature of 1.0 or 1.2. This is because lower temperatures result in less diversity of model-generating inference paths.

\begin{figure}[t!]
    \centering
    \begin{subfigure}{0.45\linewidth}
    \centering
    \includegraphics[width=\linewidth]{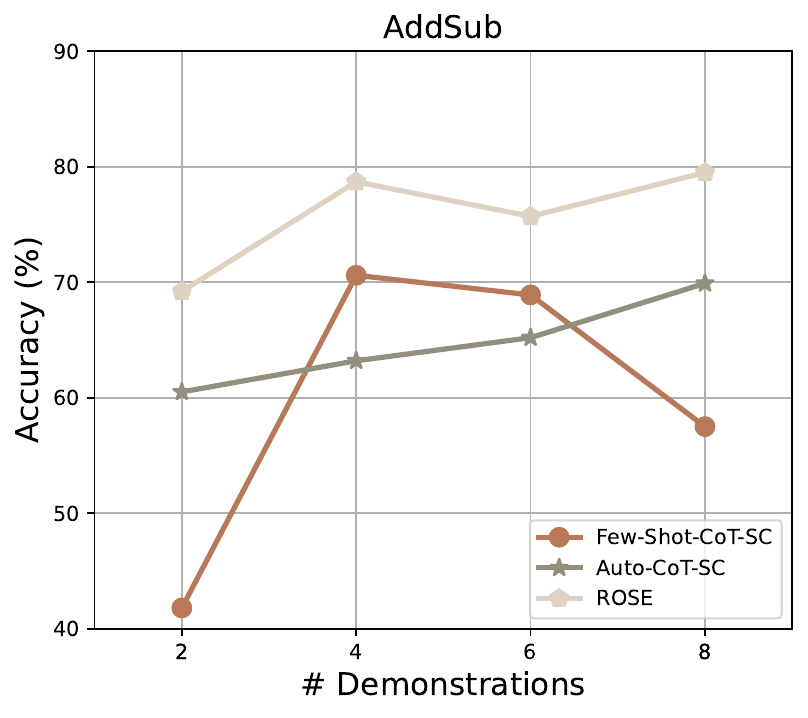}
    \end{subfigure}
    \begin{subfigure}{0.49\linewidth}
    \centering
    \includegraphics[width=\linewidth]{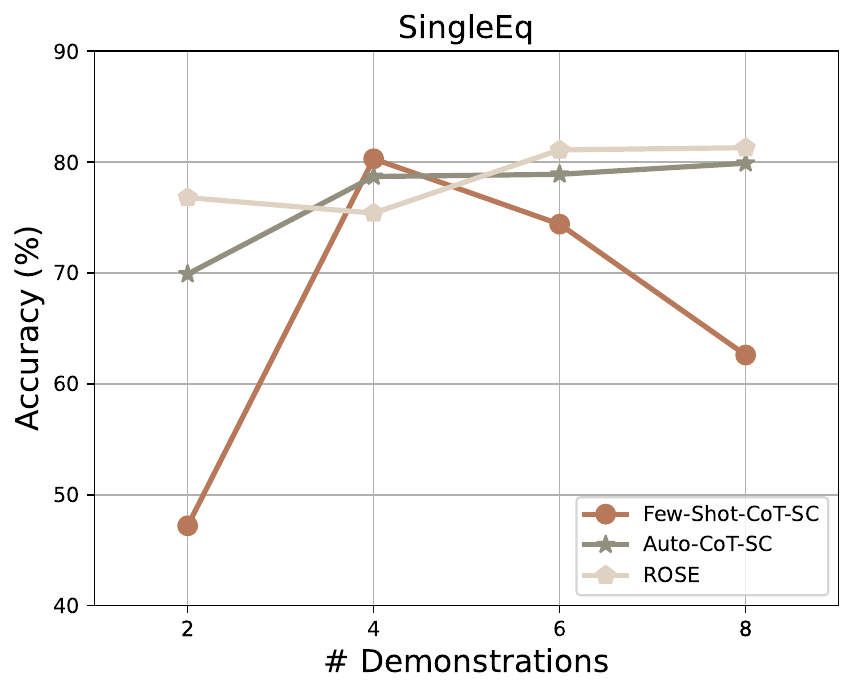}
    \end{subfigure}
    \caption{Results on different demonstration quantities. \vspace{-0.8cm}}
    \label{fig:shot}
\end{figure}

\paragraph{Results on Different Number of Reasoning Paths} Since RoSE needs to generate multiple reasoning paths for each question to estimate the uncertainty, we also evaluate RoSE under different numbers of reasoning paths. Table~\ref{tab:ablation_path} shows the results and we can see that the performance of RoSE increases with the increase of the number of reasoning paths. Moreover, RoSE consistently outperforms baseline methods across different numbers of reasoning paths, which shows the stability of RoSE.

\paragraph{Results on Different Numbers of Demonstrations} We also evaluate RoSE under different numbers of demonstrations. According to the results in Figure~\ref{fig:shot}, we see that RoSE consistently outperforms Few-Shot-CoT-SC and Auto-CoT-SC across different numbers of demonstrations, which shows the stability of RoSE. Besides, we can find that Few-Shot-CoT-SC is very unstable across different numbers of demonstrations, which also indicates that dynamically extracting demonstrations for each test question is more suitable than manual-crafting demonstrations. 

\paragraph{Transferability on Different CoT methods} RoSE is a relatively general framework that can be adapted to many CoT prompting methods. To verify the versatility of RoSE, we evaluate the performance of RoSE on two additional advanced CoT prompting methods: Plan-and-Solve~\cite{wang2023psp} and ToT~\cite{yao2023tot}. The detailed implementation settings are listed in Appendix~\ref{sec:setting_cot}.

Results on four ablation datasets are shown in Table~\ref{tab:ablation_method}. We observe that RoSE leads to consistent improvements, which shows its generality across various CoT methods. Moreover, when using the more advanced CoT methods, RoSE can get further performance improvements, which shows its potential in the future when the more powerful CoT method is proposed.

\begin{table}[ht]
\resizebox{0.98\columnwidth}{!}{\begin{tabular}{lccccc}
\toprule
\textbf{Method} & \textbf{AddSub} & \textbf{SingleEq} & \textbf{Strategy} & \textbf{Date} & \textbf{AVG}  \\
\midrule
Zero-Shot-CoT   & 83.5            & 90.9              & 65.5              & 67.5          & 76.9          \\
+ RoSE          & \textbf{90.9}   & 92.2     & \textbf{71.3}     & 88.6          & 85.8 \\
\midrule
Plan-and-Solve  & 85.6            & 91.8              & 65.9              & 68.6          & 78.0          \\
+ RoSE          & 90.6            & \textbf{94.5}              & 70.7              & \textbf{89.4} & 86.3          \\
\midrule
ToT             & 85.8            & 90.1              & 67.9              & 70.1          & 78.5          \\
+ RoSE          & 91.5            & 93.9             & \textbf{71.7}              & 88.9          & \textbf{86.5}  \\
\bottomrule
\end{tabular}}
\caption{Comparison of various CoT methods on "gpt-3.5-turbo-16k-0613" model.}
\label{tab:ablation_method}
\end{table}

\begin{figure}[ht]
    \centering
    \includegraphics[width=\linewidth]{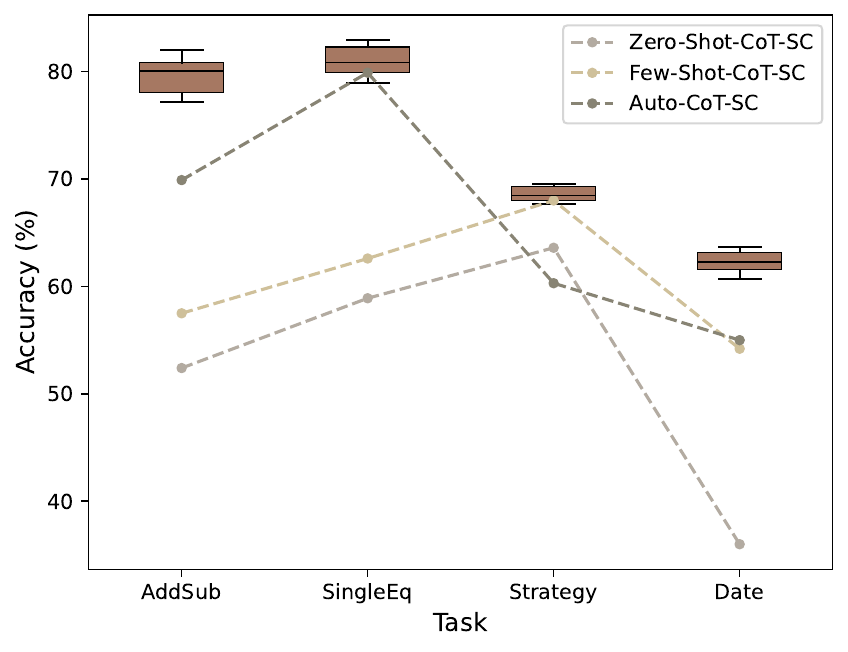}
    \caption{Results on different test orders.}
    \label{fig:ablation_consistency}
\end{figure}

\paragraph{Stability Analysis on Different Test Orders} The order of test questions will influence the performance because this can lead to different states of the experience pool. To verify the stability of RoSE, we conduct 10 evaluations on different test orders, and the distribution of results is shown in Figure~\ref{fig:ablation_consistency}. Performance fluctuates as the test order changes, but it is generally better than the baselines.

\section{Conclusion}
We present RoSE, a general framework for improving the performance of LLMS on reasoning tasks. RoSE can self-improve by constantly collecting questions into an experience pool and does not need other complex external help. To extract more helpful experience from the experience pool, we propose a systematic and novel experience orchestration mechanism that sequentially regards diversity, uncertainty, and complexity of questions in the pool as important measures to filter experiences. The comprehensive experimental results on 9 reasoning tasks and 2 LLMs show that RoSE significantly improves the reasoning performance of LLMs. Moreover, we conduct extensive analysis experiments and verify the importance of each process and the stability of RoSE across various experimental settings.

\section*{Limitations}
Since we estimate the complexity of a question using the number of reasoning steps and extract the most complex questions in the final filtering process, this may lead to a longer length of demonstrations and thus lead to slower efficiency.

\section*{Ethics Statement}
In this paper, we let LLMs self-improve on reasoning tasks. only by the collected streaming experiences. All datasets used are reasoning type and have no unsafe samples. Moreover, the LLM cannot access the internet and control external tools. Hence we think the proposed method and all experiments are safe enough, which will not cause serious impact and unrecoverable consequences on society.

\section*{Acknowledgements}
This work was supported by the National Natural Science Foundation of China (No. 62236004). The computations in this research were performed using the CFFF platform of Fudan University.

\bibliography{custom}

\newpage

\appendix

\section{Dataset Details}
\label{sec:dataset}

We evaluate RoSE on the following reasoning tasks. 
\begin{itemize}
    \item \textbf{Arithmetic reasoning.} We consider 6 Math Word Problem datasets, including AddSub~\cite{Hosseini2014addsub}, AQuA~\cite{ling2017aqua}, GSM8K~\cite{Cobbe2021gsm8k}, SingleEq~\cite{Koncel2015singleeq}, SingleOp~\cite{roy2015singleop}, and SVAMP~\cite{patel2021svamp}.
    \item \textbf{Commonsense reasoning.} We use CommonsenseQA (CSQA)~\cite{Talmor2019csqa}, StrategyQA (Strategy)~\cite{geva2021strategy}, and one dataset from BIG-bench~\cite{Srivastava2022bigbench}: Date Understanding (Date).
\end{itemize}

The detailed statistics of each task are shown in Table~\ref{tab:statistics}

\begin{table*}[ht]
    \centering
    \begin{tabular}{lccccc}
    \toprule 
    \textbf{Dataset} & \textbf{Reasoning Type} & \textbf{Answer Type} & \textbf{\# Demonstration} & \textbf{\# Test} & \textbf{License} \\
    \midrule 
    AddSub & Arithmetic & Number & 8 & 395 & Unspecified \\
    AQuA & Arithmetic & Multi-choice & 4 & 254 & Apache-2.0 \\
    GSM8K & Arithmetic & Number & 8 & 1319 & MIT License \\
    SingleEq & Arithmetic & Number & 8 & 508 & Unspecified \\
    SingleOp & Arithmetic & Number & 8 & 562 & Unspecified\\
    SVAMP & Arithmetic & Number & 8 & 1000 & MIT License \\
    \midrule
    CommonsenseQA & Commonsense & Multi-choice & 7 & 1221 & Unspecified \\
    StrategyQA & Commonsense & yes / no & 6 & 800 & MIT license \\
    Date Understanding & Commonsense & Multi-choice & 6 & 369 & MIT license \\
    \bottomrule
    \end{tabular}
    \caption{Detailed statistics of the datasets utilized in our experiment.}
    \label{tab:statistics}
\end{table*}

\section{Examples of Few Shot Methods}
\label{sec:prompt_fs}
For AddSub, AQuA, GSM8K, SingleEq, SVAMP, CommonsenseQA, and StrategyQA, we use the same few-shot demonstrations as~\citet{wei2022cot}. We manual-crafted few-shot demonstrations for other datasets. We list all demonstrations of each task for Few-Shot-CoT and Few-Shot-CoT-SC methods in Table~\ref{tab:fs_addsub} - \ref{tab:fs_date}.

\section{Implementation Details of Different CoT Methods}
\label{sec:setting_cot}
We verify the versatility of RoSE on two other CoT prompting methods: Plan-and-Solve~\cite{wang2023psp} and ToT~\cite{yao2023tot}. We also maintain a zero-shot setting for these two methods, i.e. there are no manual-crafted demonstrations. After combining the two methods with RoSE, we add each question and the corresponding thoughts into the streaming experience pool and orchestrate these collected experiences to assist in answering each new question. Although a zero-shot setting is adopted, these two methods have relatively more complex zero-shot prompts than traditional CoT methods. To take full advantage of these methods, we completed the analysis experiment on the gpt-3.5-turbo-16k-0613 model. 

For the Plan-and-Solve method, we follow the prompts in the original paper and use the same uncertainty and complexity measures as the traditional CoT method.

For ToT methods, we implement a zero-shot ToT-BFS that samples multiple thoughts using a CoT prompt and makes a vote for the best one among all thoughts. We set the step limit $T$ to 2 and generate 5 thoughts every step. To combine with our RoSE framework, we sum the percentage of the total votes for each best thought as the uncertainty measure and sum the number of steps in each best thought as the complexity measure. The prompt template for ToT is listed in Table~\ref{tab:tot_prompt}

\begin{table*}[ht]
    \centering
    \begin{tabularx}{\textwidth}{X}
    \toprule 
    \textbf{Q: }There are 15 trees in the grove. Grove workers will plant trees in the grove today. After they are done, there will be 21 trees. How many trees did the grove workers plant today? \\
    \textbf{A: } Let's think step by step. There are originally 3 cars. 2 more cars arrive. 3 + 2 = 5. The answer is 5. \\
    \hdashline 
    \textbf{Q: }If there are 3 cars in the parking lot and 2 more cars arrive, how many cars are in the parking lot? \\
    \textbf{A: } Let's think step by step. There are 15 trees originally. Then there were 21 trees after some more were planted. So there must have been 21 - 15 = 6. The answer is 6.\\
    \hdashline
    \textbf{Q: }Leah had 32 chocolates and her sister had 42. If they ate 35, how many pieces do they have left in total? \\
    \textbf{A: }Let's think step by step. Originally, Leah had 32 chocolates. Her sister had 42. So in total, they had 32 + 42 = 74. After eating 35, they had 74 - 35 = 39. The answer is 39.\\
    \hdashline
    \textbf{Q: }Jason had 20 lollipops. He gave Denny some lollipops. Now Jason has 12 lollipops. How many lollipops did Jason give to Denny? \\
    \textbf{A: }Let's think step by step. Jason started with 20 lollipops. Then he had 12 after giving some to Denny. So he gave Denny 20 - 12 = 8. The answer is 8.\\
    \hdashline
    \textbf{Q: }Shawn has five toys. For Christmas, he got two toys each from his mom and dad. How many toys does he have now? \\
    \textbf{A: }Let's think step by step. There are 15 trees originally. Shawn started with 5 toys. If he got 2 toys each from his mom and dad, then that is 4 more toys. 5 + 4 = 9. The answer is 9.\\
    \hdashline
    \textbf{Q: }There were nine computers in the server room. Five more computers were installed each day, from Monday to Thursday. How many computers are now in the server room? \\
    \textbf{A: }Let's think step by step. There were originally 9 computers. For each of 4 days, 5 more computers were added. So 5 * 4 = 20 computers were added. 9 + 20 is 29. The answer is 29.\\
    \hdashline
    \textbf{Q: }Michael had 58 golf balls. On Tuesday, he lost 23 golf balls. On Wednesday, he lost 2 more. How many golf balls did he have at the end of Wednesday? \\
    \textbf{A: }Let's think step by step. There are 15 trees originally. Michael started with 58 golf balls. After losing 23 on Tuesday, he had 58 - 23 = 35. After losing 2 more, he had 35 - 2 = 33 golf balls. The answer is 33.\\
    \hdashline
    \textbf{Q: }Olivia has \$23. She bought five bagels for \$3 each. How much money does she have left? \\
    \textbf{A: }Let's think step by step. Olivia had 23 dollars. 5 bagels for 3 dollars each will be 5 x 3 = 15 dollars. So she has 23 - 15 dollars left. 23 - 15 is 8. The answer is 8.\\
    \bottomrule
    \end{tabularx}
    \caption{Few-Shot Demonstrations for AddSub.}
    \label{tab:fs_addsub}
\end{table*}

\begin{table*}[ht]
    \centering
    \begin{tabularx}{\textwidth}{X}
    \toprule 
    \textbf{Q: }John found that the average of 15 numbers is 40. If 10 is added to each number then the mean of the numbers is? \\
    Answer Choices: (A) 50 (B) 45 (C) 65 (D) 78 (E) 64 \\
    \textbf{A: } Let's think step by step. If 10 is added to each number, then the mean of the numbers also increases by 10. So the new mean would be 50. The answer is A. \\
    \hdashline 
    \textbf{Q: }If a / b = 3/4 and 8a + 5b = 22, then find the value of a. \\
    Answer Choices: (A) 1/2 (B) 3/2 (C) 5/2 (S) 4/2 (E) 7/2 \\
    \textbf{A: } Let's think step by step. If a / b = 3/4, then b = 4a / 3. So 8a + 5(4a / 3) = 22. This simplifies to 8a + 20a / 3 = 22, which means 44a / 3 = 22. So a is equal to 3/2. The answer is B.\\
    \hdashline
    \textbf{Q: }A person is traveling at 20 km/hr and reached his destiny in 2.5 hr then find the distance? \\
    Answer Choices: (A) 53 km (B) 55 km (C) 52 km (D) 60 km (E) 50 km \\
    \textbf{A: }Let's think step by step. The distance that the person traveled would have been 20 km/hr * 2.5 hrs = 50 km. The answer is E.\\
    \hdashline
    \textbf{Q: }How many keystrokes are needed to type the numbers from 1 to 500? \\
    Answer Choices: (A) 1156 (B) 1392 (C) 1480 (D) 1562 (E) 1788 \\
    \textbf{A: }Let's think step by step. There are 9 one-digit numbers from 1 to 9. There are 90 two-digit numbers from 10 to 99. There are 401 three-digit numbers from 100 to 500. 9 + 90(2) + 401(3) = 1392. The answer is B.\\
    \bottomrule
    \end{tabularx}
    \caption{Few-Shot Demonstrations for AQuA.}
    \label{tab:fs_aqua}
\end{table*}

\begin{table*}[ht]
    \centering
    \begin{tabularx}{\textwidth}{X}
    \toprule 
    \textbf{Q: }There are 15 trees in the grove. Grove workers will plant trees in the grove today. After they are done, there will be 21 trees. How many trees did the grove workers plant today? \\
    \textbf{A: } Let's think step by step. There are originally 3 cars. 2 more cars arrive. 3 + 2 = 5. The answer is 5. \\
    \hdashline 
    \textbf{Q: }If there are 3 cars in the parking lot and 2 more cars arrive, how many cars are in the parking lot? \\
    \textbf{A: } Let's think step by step. There are 15 trees originally. Then there were 21 trees after some more were planted. So there must have been 21 - 15 = 6. The answer is 6.\\
    \hdashline
    \textbf{Q: }Leah had 32 chocolates and her sister had 42. If they ate 35, how many pieces do they have left in total? \\
    \textbf{A: }Let's think step by step. Originally, Leah had 32 chocolates. Her sister had 42. So in total they had 32 + 42 = 74. After eating 35, they had 74 - 35 = 39. The answer is 39.\\
    \hdashline
    \textbf{Q: }Jason had 20 lollipops. He gave Denny some lollipops. Now Jason has 12 lollipops. How many lollipops did Jason give to Denny? \\
    \textbf{A: }Let's think step by step. Jason started with 20 lollipops. Then he had 12 after giving some to Denny. So he gave Denny 20 - 12 = 8. The answer is 8.\\
    \hdashline
    \textbf{Q: }Shawn has five toys. For Christmas, he got two toys each from his mom and dad. How many toys does he have now? \\
    \textbf{A: }Let's think step by step. There are 15 trees originally. Shawn started with 5 toys. If he got 2 toys each from his mom and dad, then that is 4 more toys. 5 + 4 = 9. The answer is 9.\\
    \hdashline
    \textbf{Q: }There were nine computers in the server room. Five more computers were installed each day, from monday to thursday. How many computers are now in the server room? \\
    \textbf{A: }Let's think step by step. There were originally 9 computers. For each of 4 days, 5 more computers were added. So 5 * 4 = 20 computers were added. 9 + 20 is 29. The answer is 29.\\
    \hdashline
    \textbf{Q: }Michael had 58 golf balls. On tuesday, he lost 23 golf balls. On wednesday, he lost 2 more. How many golf balls did he have at the end of wednesday? \\
    \textbf{A: }Let's think step by step. There are 15 trees originally. Michael started with 58 golf balls. After losing 23 on tuesday, he had 58 - 23 = 35. After losing 2 more, he had 35 - 2 = 33 golf balls. The answer is 33.\\
    \hdashline
    \textbf{Q: }Olivia has \$23. She bought five bagels for \$3 each. How much money does she have left? \\
    \textbf{A: }Let's think step by step. Olivia had 23 dollars. 5 bagels for 3 dollars each will be 5 x 3 = 15 dollars. So she has 23 - 15 dollars left. 23 - 15 is 8. The answer is 8.\\
    \bottomrule
    \end{tabularx}
    \caption{Few-Shot Demonstrations for GSM8K.}
    \label{tab:fs_gsm8k}
\end{table*}

\begin{table*}[ht]
    \centering
    \begin{tabularx}{\textwidth}{X}
    \toprule 
    \textbf{Q: }There are 15 trees in the grove. Grove workers will plant trees in the grove today. After they are done, there will be 21 trees. How many trees did the grove workers plant today? \\
    \textbf{A: } Let's think step by step. There are originally 3 cars. 2 more cars arrive. 3 + 2 = 5. The answer is 5. \\
    \hdashline 
    \textbf{Q: }If there are 3 cars in the parking lot and 2 more cars arrive, how many cars are in the parking lot? \\
    \textbf{A: } Let's think step by step. There are 15 trees originally. Then there were 21 trees after some more were planted. So there must have been 21 - 15 = 6. The answer is 6.\\
    \hdashline
    \textbf{Q: }Leah had 32 chocolates and her sister had 42. If they ate 35, how many pieces do they have left in total? \\
    \textbf{A: }Let's think step by step. Originally, Leah had 32 chocolates. Her sister had 42. So in total they had 32 + 42 = 74. After eating 35, they had 74 - 35 = 39. The answer is 39.\\
    \hdashline
    \textbf{Q: }Jason had 20 lollipops. He gave Denny some lollipops. Now Jason has 12 lollipops. How many lollipops did Jason give to Denny? \\
    \textbf{A: }Let's think step by step. Jason started with 20 lollipops. Then he had 12 after giving some to Denny. So he gave Denny 20 - 12 = 8. The answer is 8.\\
    \hdashline
    \textbf{Q: }Shawn has five toys. For Christmas, he got two toys each from his mom and dad. How many toys does he have now? \\
    \textbf{A: }Let's think step by step. There are 15 trees originally. Shawn started with 5 toys. If he got 2 toys each from his mom and dad, then that is 4 more toys. 5 + 4 = 9. The answer is 9.\\
    \hdashline
    \textbf{Q: }There were nine computers in the server room. Five more computers were installed each day, from monday to thursday. How many computers are now in the server room? \\
    \textbf{A: }Let's think step by step. There were originally 9 computers. For each of 4 days, 5 more computers were added. So 5 * 4 = 20 computers were added. 9 + 20 is 29. The answer is 29.\\
    \hdashline
    \textbf{Q: }Michael had 58 golf balls. On tuesday, he lost 23 golf balls. On wednesday, he lost 2 more. How many golf balls did he have at the end of wednesday? \\
    \textbf{A: }Let's think step by step. There are 15 trees originally. Michael started with 58 golf balls. After losing 23 on tuesday, he had 58 - 23 = 35. After losing 2 more, he had 35 - 2 = 33 golf balls. The answer is 33.\\
    \hdashline
    \textbf{Q: }Olivia has \$23. She bought five bagels for \$3 each. How much money does she have left? \\
    \textbf{A: }Let's think step by step. Olivia had 23 dollars. 5 bagels for 3 dollars each will be 5 x 3 = 15 dollars. So she has 23 - 15 dollars left. 23 - 15 is 8. The answer is 8.\\
    \bottomrule
    \end{tabularx}
    \caption{Few-Shot Demonstrations for SingleEq.}
    \label{tab:fs_singleeq}
\end{table*}

\begin{table*}[ht]
    \centering
    \begin{tabularx}{\textwidth}{X}
    \toprule 
    \textbf{Q: }There are 15 trees in the grove. Grove workers will plant trees in the grove today. After they are done, there will be 21 trees. How many trees did the grove workers plant today? \\
    \textbf{A: } Let's think step by step. There are originally 3 cars. 2 more cars arrive. 3 + 2 = 5. The answer is 5. \\
    \hdashline 
    \textbf{Q: }If there are 3 cars in the parking lot and 2 more cars arrive, how many cars are in the parking lot? \\
    \textbf{A: } Let's think step by step. There are 15 trees originally. Then there were 21 trees after some more were planted. So there must have been 21 - 15 = 6. The answer is 6.\\
    \hdashline
    \textbf{Q: }Leah had 32 chocolates and her sister had 42. If they ate 35, how many pieces do they have left in total? \\
    \textbf{A: }Let's think step by step. Originally, Leah had 32 chocolates. Her sister had 42. So in total they had 32 + 42 = 74. After eating 35, they had 74 - 35 = 39. The answer is 39.\\
    \hdashline
    \textbf{Q: }Jason had 20 lollipops. He gave Denny some lollipops. Now Jason has 12 lollipops. How many lollipops did Jason give to Denny? \\
    \textbf{A: }Let's think step by step. Jason started with 20 lollipops. Then he had 12 after giving some to Denny. So he gave Denny 20 - 12 = 8. The answer is 8.\\
    \hdashline
    \textbf{Q: }Shawn has five toys. For Christmas, he got two toys each from his mom and dad. How many toys does he have now? \\
    \textbf{A: }Let's think step by step. There are 15 trees originally. Shawn started with 5 toys. If he got 2 toys each from his mom and dad, then that is 4 more toys. 5 + 4 = 9. The answer is 9.\\
    \hdashline
    \textbf{Q: }There were nine computers in the server room. Five more computers were installed each day, from monday to thursday. How many computers are now in the server room? \\
    \textbf{A: }Let's think step by step. There were originally 9 computers. For each of 4 days, 5 more computers were added. So 5 * 4 = 20 computers were added. 9 + 20 is 29. The answer is 29.\\
    \hdashline
    \textbf{Q: }Michael had 58 golf balls. On tuesday, he lost 23 golf balls. On wednesday, he lost 2 more. How many golf balls did he have at the end of wednesday? \\
    \textbf{A: }Let's think step by step. There are 15 trees originally. Michael started with 58 golf balls. After losing 23 on tuesday, he had 58 - 23 = 35. After losing 2 more, he had 35 - 2 = 33 golf balls. The answer is 33.\\
    \hdashline
    \textbf{Q: }Olivia has \$23. She bought five bagels for \$3 each. How much money does she have left? \\
    \textbf{A: }Let's think step by step. Olivia had 23 dollars. 5 bagels for 3 dollars each will be 5 x 3 = 15 dollars. So she has 23 - 15 dollars left. 23 - 15 is 8. The answer is 8.\\
    \bottomrule
    \end{tabularx}
    \caption{Few-Shot Demonstrations for SingleOp.}
    \label{tab:fs_singleop}
\end{table*}

\begin{table*}[ht]
    \centering
    \begin{tabularx}{\textwidth}{X}
    \toprule 
    \textbf{Q: }There are 15 trees in the grove. Grove workers will plant trees in the grove today. After they are done, there will be 21 trees. How many trees did the grove workers plant today? \\
    \textbf{A: } Let's think step by step. There are originally 3 cars. 2 more cars arrive. 3 + 2 = 5. The answer is 5. \\
    \hdashline 
    \textbf{Q: }If there are 3 cars in the parking lot and 2 more cars arrive, how many cars are in the parking lot? \\
    \textbf{A: } Let's think step by step. There are 15 trees originally. Then there were 21 trees after some more were planted. So there must have been 21 - 15 = 6. The answer is 6.\\
    \hdashline
    \textbf{Q: }Leah had 32 chocolates and her sister had 42. If they ate 35, how many pieces do they have left in total? \\
    \textbf{A: }Let's think step by step. Originally, Leah had 32 chocolates. Her sister had 42. So in total they had 32 + 42 = 74. After eating 35, they had 74 - 35 = 39. The answer is 39.\\
    \hdashline
    \textbf{Q: }Jason had 20 lollipops. He gave Denny some lollipops. Now Jason has 12 lollipops. How many lollipops did Jason give to Denny? \\
    \textbf{A: }Let's think step by step. Jason started with 20 lollipops. Then he had 12 after giving some to Denny. So he gave Denny 20 - 12 = 8. The answer is 8.\\
    \hdashline
    \textbf{Q: }Shawn has five toys. For Christmas, he got two toys each from his mom and dad. How many toys does he have now? \\
    \textbf{A: }Let's think step by step. There are 15 trees originally. Shawn started with 5 toys. If he got 2 toys each from his mom and dad, then that is 4 more toys. 5 + 4 = 9. The answer is 9.\\
    \hdashline
    \textbf{Q: }There were nine computers in the server room. Five more computers were installed each day, from monday to thursday. How many computers are now in the server room? \\
    \textbf{A: }Let's think step by step. There were originally 9 computers. For each of 4 days, 5 more computers were added. So 5 * 4 = 20 computers were added. 9 + 20 is 29. The answer is 29.\\
    \hdashline
    \textbf{Q: }Michael had 58 golf balls. On tuesday, he lost 23 golf balls. On wednesday, he lost 2 more. How many golf balls did he have at the end of wednesday? \\
    \textbf{A: }Let's think step by step. There are 15 trees originally. Michael started with 58 golf balls. After losing 23 on tuesday, he had 58 - 23 = 35. After losing 2 more, he had 35 - 2 = 33 golf balls. The answer is 33.\\
    \hdashline
    \textbf{Q: }Olivia has \$23. She bought five bagels for \$3 each. How much money does she have left? \\
    \textbf{A: }Let's think step by step. Olivia had 23 dollars. 5 bagels for 3 dollars each will be 5 x 3 = 15 dollars. So she has 23 - 15 dollars left. 23 - 15 is 8. The answer is 8.\\
    \bottomrule
    \end{tabularx}
    \caption{Few-Shot Demonstrations for SVAMP.}
    \label{tab:fs_svamp}
\end{table*}

\begin{table*}[ht]
    \centering
    \begin{tabularx}{\textwidth}{X}
    \toprule 
    \textbf{Q: }What do people use to absorb extra ink from a fountain pen? \\
    Answer Choices: (A) shirt pocket (B) calligrapher’s hand (C) inkwell (D) desk drawer (E) blotter \\
    \textbf{A: } Let's think step by step. The answer must be an item that can absorb ink. Of the above choices, only blotters are used to absorb ink. The answer is E. \\
    \hdashline 
    \textbf{Q: }What home entertainment equipment requires cable? \\
    Answer Choices: (A) radio shack (B) substation (C) television (D) cabinet \\
    \textbf{A: } Let's think step by step. The answer must require cable. Of the above choices, only television requires cable. The answer is C.\\
    \hdashline
    \textbf{Q: }The fox walked from the city into the forest, what was it looking for? \\
    Answer Choices: (A) pretty flowers (B)hen house (C) natural habitat (D) storybook \\
    \textbf{A: }Let's think step by step. The answer must be something in the forest. Of the above choices, only natural habitat is in the forest. The answer is C.\\
    \hdashline
    \textbf{Q: }Sammy wanted to go to where the people were. Where might he go? \\
    Answer Choices: (A) populated areas (B) race track (C) desert (D) apartment (E) roadblock \\
    \textbf{A: }Let's think step by step. The answer must be a place with a lot of people. Of the above choices, only populated areas have a lot of people. The answer is A.\\
    \hdashline
    \textbf{Q: }Where do you put your grapes just before checking out? \\
    Answer Choices: (A) mouth (B) grocery cart (C)super market (D) fruit basket (E) fruit market \\
    \textbf{A: }Let's think step by step. The answer should be the place where grocery items are placed before checking out. Of the above choices, grocery cart makes the most sense for holding grocery items. The answer is B.\\
    \hdashline
    \textbf{Q: }Google Maps and other highway and street GPS services have replaced what? \\
    Answer Choices: (A) united states (B) mexico (C) countryside (D) atlas \\
    \textbf{A: }Let's think step by step. The answer must be something that used to do what Google Maps and GPS services do, which is to give directions. Of the above choices, only atlases are used to give directions. The answer is D.\\
    \hdashline
    \textbf{Q: }Before getting a divorce, what did the wife feel who was doing all the work? \\
    Answer Choices: (A) harder (B) anguish (C) bitterness (D) tears (E) sadness \\
    \textbf{A: }Let's think step by step. The answer should be the feeling of someone getting divorced who was doing all the work. Of the above choices, the closest feeling is bitterness. The answer is C.\\
    \bottomrule
    \end{tabularx}
    \caption{Few-Shot Demonstrations for CommonsenseQA.}
    \label{tab:fs_csqa}
\end{table*}

\begin{table*}[ht]
    \centering
    \begin{tabularx}{\textwidth}{X}
    \toprule 
    \textbf{Q: }Do hamsters provide food for any animals? \\
    \textbf{A: } Let's think step by step. Hamsters are prey animals. Prey are food for predators. Thus, hamsters provide food for some animals. The answer is yes. \\
    \hdashline 
    \textbf{Q: }Could Brooke Shields succeed at University of Pennsylvania? \\
    \textbf{A: } Let's think step by step. Brooke Shields went to Princeton University. Princeton University is about as academically rigorous as the University of Pennsylvania. Thus, Brooke Shields could also succeed at the University of Pennsylvania. The answer is yes.\\
    \hdashline
    \textbf{Q: }Yes or no: Hydrogen’s atomic number squared exceeds number of Spice Girls? \\
    \textbf{A: }Let's think step by step. Hydrogen has an atomic number of 1. 1 squared is 1. There are 5 Spice Girls. Thus, Hydrogen’s atomic number squared is less than 5. The answer is no.\\
    \hdashline
    \textbf{Q: }Yes or no: Is it common to see frost during some college commencements? \\
    \textbf{A: }Let's think step by step. College commencement ceremonies can happen in December, May, and June. December is in the winter, so there can be frost. Thus, there could be frost at some commencements. The answer is yes.\\
    \hdashline
    \textbf{Q: }Yes or no: Could a llama birth twice during War in Vietnam (1945-46)? \\
    \textbf{A: }Let's think step by step. The War in Vietnam was 6 months. The gestation period for a llama is 11 months, which is more than 6 months. Thus, a llama could not give birth twice during the War in Vietnam. The answer is no.\\
    \hdashline
    \textbf{Q: }Yes or no: Would a pear sink in water? \\
    \textbf{A: }Let's think step by step. The density of a pear is about 0.6g/cm3, which is less than water. Objects less dense than water float. Thus, a pear would float. The answer is no.\\
    \bottomrule
    \end{tabularx}
    \caption{Few-Shot Demonstrations for StrategyQA.}
    \label{tab:fs_Strategy}
\end{table*}

\begin{table*}[ht]
    \centering
    \begin{tabularx}{\textwidth}{X}
    \toprule 
    \textbf{Q: }2015 is coming in 36 hours. What is the date one week from today in MM/DD/YYYY? \\
    Answer Choices: (A) 01/05/2015 (B) 01/06/2015 (C) 01/04/2015 (D) 02/05/2015 (E) 12/05/2015 (F) 01/05/2016 \\
    \textbf{A: } Let's think step by step. If 2015 is coming in 36 hours, then it is coming in 2 days. 2 days before 01/01/2015 is 12/30/2014, so today is 12/30/2014. So one week from today will be 01/05/2015. The answer is A. \\
    \hdashline 
    \textbf{Q: }The first day of 2019 is a Tuesday, and today is the first Monday of 2019. What is the date today in MM/DD/YYYY? \\
    Answer Choices: (A) 01/08/2019 (B) 01/07/2019 (C) 01/06/2019 (D) 02/07/2019 (E) 12/07/2019 (F) 01/07/2018 \\
    \textbf{A: } Let's think step by step. If the first day of 2019 was Tuesday, then 01/01/2019 was a Tuesday. Today is the first monday, would be six days later. So today is 01/07/2019. The answer is B.\\
    \hdashline
    \textbf{Q: }The concert was scheduled to be on 06/01/1943, but was delayed by one day to today. What is the date 10 days ago in MM/DD/YYYY? \\
    Answer Choices: (A) 05/22/1943 (B) 05/23/1943 (C) 05/24/1943 (D) 05/25/1943 (E) 05/26/1943 (F) 05/27/1943 \\
    \textbf{A: }Let's think step by step. One day after 06/01/1943 is 06/02/1943, so today is 06/02/1943. 10 days before today is 05/23/1943. The answer is B.\\
    \hdashline
    \textbf{Q: }It is 4/19/1969 today. What is the date 24 hours later in MM/DD/YYYY? \\ Answer Choices: (A) 04/23/1969 (B) 04/21/1969 (C) 04/22/1969 (D) 04/20/1969 (E) 04/24/1969 (F) 04/25/1969 \\
    \textbf{A: }Let's think step by step. Today is 04/19/1969. 24 hours later is one day after today, which would be 04/20/1969. The answer is D.\\
    \hdashline
    \textbf{Q: }Jane thought today is 3/11/2002, but today is in fact Mar 12, which is 1 day later. What is the date 24 hours later in MM/DD/YYYY? \\
    Answer Choices: (A) 03/17/2002 (B) 03/14/2002 (C) 03/15/2002 (D) 03/16/2002 (E) 03/13/2002 (F) 03/18/2002 \\
    \textbf{A: }Let's think step by step. Today is 03/12/2002. So the date 24 hours later will be 03/13/2002. The answer is E.\\
    \hdashline
    \textbf{Q: }Jane was born on the last day of Feburary in 2001. Today is her 16-year-old birthday. What is the date yesterday in MM/DD/YYYY? \\
    Answer Choices: (A) 03/04/2017 (B) 02/28/2017 (C) 03/01/2017 (D) 03/02/2017 (E) 03/03/2017 (F) 02/27/2017 \\
    \textbf{A: }Let's think step by step. The last day of February is the 28th, so Jane was born on 02/28/2001. Today is her 16-year old birthday.So yesterday was 02/27/2017. The answer is F.\\
    \bottomrule
    \end{tabularx}
    \caption{Few-Shot Demonstrations for Date Understanding.}
    \label{tab:fs_date}
\end{table*}

\begin{table*}[ht]
    \centering
    \begin{tabularx}{\textwidth}{l}
    \toprule 
    \underline{\textbf{Answer Format}} \\
    addsub\_format ~=~ '"the answer is n" where n is a number' \\
    single\_format ~=~ '"the answer is n" where n is a number' \\
    strategy\_format ~=~ 'either "the answer is yes" or "the answer is no"' \\
    date\_format ~=~ '"the answer is n" where n is one of "A, B, C, D, E, F"' \\
    \underline{\textbf{Thought Format}} \\
    Answer the following question: \{input\} \\
    Make a strategy then write. Your output should be of the following format: \\
    Strategy: \\
    Your strategy about how to answer the question. \\
    Answer: \\
    Your answer to the question. It should end with \{format\}. \\
    \underline{\textbf{Voting Prompt}} \\
    Given an instruction and several choices, decide which choice is most promising. \\
    Analyze each choice in detail, then conclude in the last line \\"The best choice is \{s\}", where s is the integer id of the choice. \\
    \bottomrule
    \end{tabularx}
    \caption{Prompt template for ToT methods.}
    \label{tab:tot_prompt}
\end{table*}

\end{document}